\documentclass[runningheads]{llncs}
\usepackage{graphicx}
\usepackage{mdframed}
\usepackage{booktabs}
\begin{document}
\title{MORTY: Structured Summarization for Targeted Information Extraction from Scholarly Articles}
\titlerunning{Targeted Information Extraction from Scholarly Articles}
%
\author{Mohamad Yaser Jaradeh\inst{1}\orcidID{0000-0001-8777-2780} \and
Markus Stocker\inst{2}\orcidID{0000-0001-5492-3212} \and
Sören Auer\inst{2}\orcidID{0000-0002-0698-2864}}
\authorrunning{Jaradeh et al.}
%
\institute{L3S Research Center, Leibniz University Hannover, Germany \\
\email{jaradeh@l3s.de} \and
Leibniz Information Centre for Science and Technology \\
\email{\{auer,markus.stocker\}@tib.eu}}
\maketitle              
\begin{abstract}
Information extraction from scholarly articles is a challenging task due to the sizable document length and implicit information hidden in text, figures, and citations.
Scholarly information extraction has various applications in exploration, archival, and curation services for digital libraries and knowledge management systems.
We present \textsc{MORTY}, an information extraction technique that creates structured summaries of text from scholarly articles.
Our approach condenses the article's full-text to property-value pairs as a segmented text snippet called structured summary.
We also present a sizable scholarly dataset combining structured summaries retrieved from a scholarly knowledge graph and corresponding publicly available scientific articles, which we openly publish as a resource for the research community.
Our results show that structured summarization is a suitable approach for targeted information extraction that complements other commonly used methods such as question answering and named entity recognition.

\end{abstract}

\section{Introduction}
\label{sec:intro}

By their very nature, scholarly articles tend to be dense with information and knowledge~\cite{Palmatier2018Review}.
The task of information extraction (IE) has been widely researched by the community in a variety of contexts~\cite{chang2006survey,piskorski2013information,jiang2012information}, including the scholarly domain~\cite{williams2016information,nasar2018information}.
However, information extraction from scholarly articles continues to suffer from low accuracy.
Reasons include ambiguity of scholarly text, information representation in scholarly articles, and lack of training datasets~\cite{Singh2016OCR}.

Other than retrospective information extraction, initiatives such as the ORKG~\cite{jaradeh2019open}, Hi-Knowledge~\cite{Jeschke2020hi}, and Coda~\cite{Spadaro2020The} collect structured scholarly information by engaging researchers in the knowledge curation process.
In ORKG, information is collected by experts that extract and structure the essential information from articles. 
However, experts might not use the exact wording from the original article or might put forward a novel segment of text that did not exist before in the original text.

Information extraction techniques~\cite{sarawagi2008information} could play a supporting role through automated extraction, suggestions to experts or autonomously adding extracted information to a a data source (e.g. database or knowledge graph).
However, blindly extracting information (i.e. factual extractions) is not suitable for scholarly data due to the large amount of information condensed into little text.
Blind extraction refers to Open Information Extraction~\cite{etzioni2008open} that relies on propositions and facts as well as common entities and relations between them.
For scholarly articles, a more targeted approach is required, whereby a system is able to extract a set of predefined properties and their corresponding values while ignoring others.

We propose \textsc{MORTY}, a method that leverages summarization tasks conducted by deep-learning language models to create structured summaries that can be parsed into extracted information, stored in a knowledge graph.
We present, evaluate, and discuss \textsc{MORTY}. Furthermore, we highlight the research problems, possible solutions, limitations of the approach, and review open questions and future prospects.

The core contributions of this article are: First, a dataset of paper full texts with a list of property-value pairs of human-expert annotated information.
Second, an approach for information extraction from scholarly articles using structured summarization.
\begin{figure}[t]
  \centering
  \includegraphics[width=\textwidth]{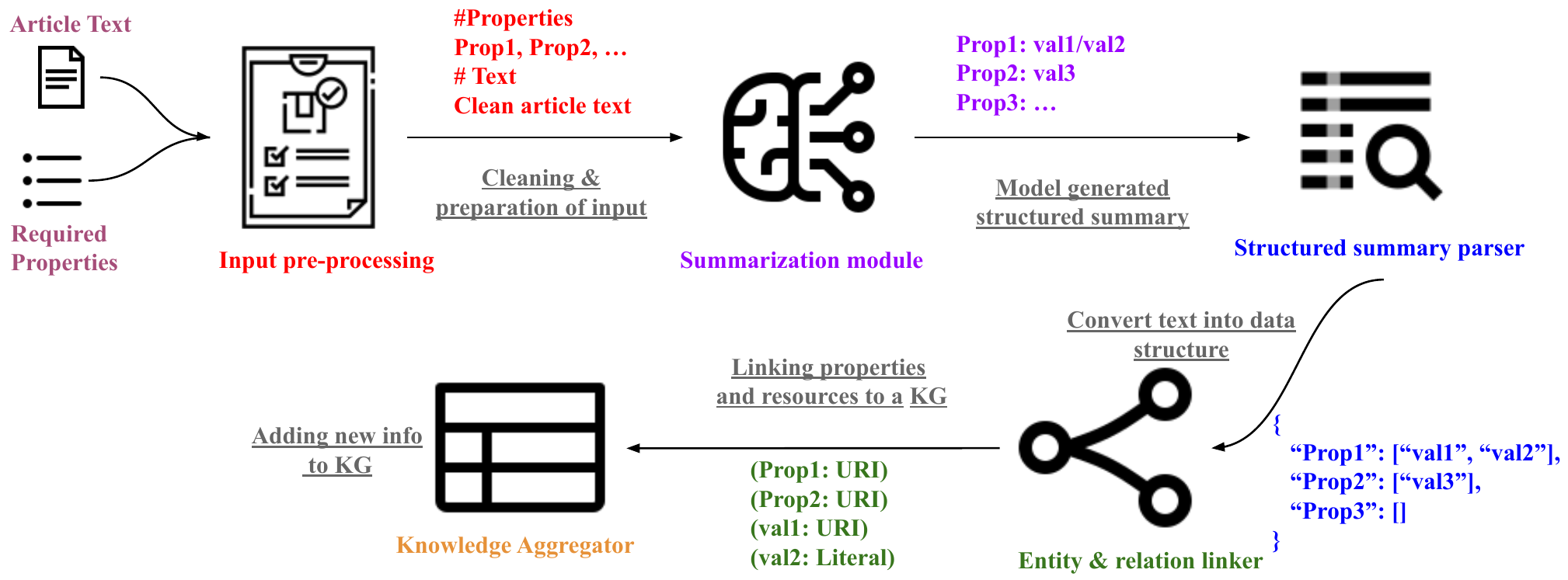}
  \caption{Bird's eye view on the complete workflow of employing structured summarization in the context of information extraction from scholarly documents and articles (MORTY).}
  \label{fig:approach}
\end{figure}

\section{Related Work}
\label{sec:rw}

\textbf{Information Extraction.}
Several information extraction methods have been proposed by the community, each with their own advantages and disadvantages.
Ji et al.~\cite{JI2020An} proposed an end-to-end system that uses a multi-task model to perform sentence classification and information extraction on legal documents.
TRIE~\cite{Zhang2020TRIE} uses end-to-end system to jointly perform document reading and information extraction on everyday documents such as invoices, tickets, and resumes.
Chua and Duffy~ \cite{Chua2021DeepCPCFG} proposes a method for finding the suitable grammar set for the parsing and the extraction of information. 
Specifically for scholarly context, various systems has been created to extract and retrieve information from publications and scholarly articles.
exBERT~\cite{Jaradeh2021Triple} uses triple classification to perform knowledge graph completion.
Dasigi et al.~\cite{dasigi2021dataset} proposed a method to retrieve information from papers to answer natural language questions.
FNG-IE~\cite{Tahir2021FNGIEAI} is an improved graph-based approach for the extraction of keywords from scholarly big-data.
Furthermore, Liu et al.~\cite{Liu2007TableSeer} presents the TableSeer system that is capable of metadata extraction from tables of scholarly nature.

\textbf{Language Models.}
With regards to automated text summarization, various language models relying on attention mechanisms~\cite{Vaswani2017Attention} displayed state-of-the-art results superseding human performance.
BERT~\cite{devlin2018bert} (scholarly counterpart SciBERT~\cite{beltagy2019scibert}) are some of the most commonly used transformer models capable to automatically summarize text.
Similarly, RoBERTa~\cite{liu2019roberta} is an optimized approach to represent language and is capable of producing summarizations of text.
BART~\cite{lewis2019bart} is a sequence-to-sequence model trained as a denoising autoencoder, which improves on the pre-training phase.
Zhang et al.~\cite{zhang2020pegasus} presents the PEGASUS model trained for abstractive summary generation of text.
These and other models are usually built to handle ``short'' input sequences, e.g. 512-1K tokens.
Other attempts address the issue of processing longer inputs.
BigBird~\cite{zaheer2020big} and Longformer~\cite{beltagy2020longformer} present models that are capable of handling a much larger input, e.g. 4K-16K tokens.
Other generational models aren't created specifically for one task; instead, they are capable of performing multiple tasks depending on the input text.
For instance, GPT2~\cite{radford2019language} supports unconditional text generation.
Raffel et al.~\cite{raffel2019exploring} describe T5, a model that can perform summarization, translation, and question answering based on keywords in the input text.
Some of these language models have been either pre-trained on scholarly data such as PubMed\footnote{\url{https://www.nlm.nih.gov/databases/download/pubmed\_medline.html}} and arXiv~\cite{clement2019arxiv} datasets, or have been fine-tuned on such data for empirical evaluation in their original publication.
Our contribution in this article leverages the capabilities of automatic summarization for the objective of information extraction form scholarly documents.

\section{MORTY}
\label{sec:morty}
Scholarly text is ambiguous and information dense.
We illustrate the problem by taking a look at the abstract of this article.
If we want to extract a single piece of information (i.e. a property) such as the ``research problem'' addressed by the article, it is necessary to comprehend the text and look even behind the textual representation.
In this example, the research problem is ``information extraction from scholarly articles''.
The method of looking up certain properties such as ``research problem'' in the text, proves insufficient because the phrase may not exist as is or is spelled differently.
This can be extend by looking up synonyms for the property or by finding verbs that represent the same intent (e.g., addresses, tackles, etc.).
Other times, regardless of how the property is represented, the value itself is implicit or not represented as expected, which requires more abstractive answers than extractive ones~\cite{tas2007survey}.

We argue that these cases barely scratch the surface of the problem.
Certain properties could require values placed throughout the text, combined together, and even morphed into dissimilar wording.
Others, cannot be found in the text, but are included in figures, tables, or even in citations~\cite{ray2015automatic,Xia2017Big}.
Furthermore, some properties could be of annotation-nature, i.e. the property and the value are not in the original text, but tacit knowledge of an expert annotating an article.




\textsc{MORTY} leverages the capabilities of deep learning language models to comprehend the semantics of scholarly text and perform targeted information extraction via text summarization.
Scholarly articles typically follow a certain structure. IMRaD~\cite{sollaci2004IMRaD} refers to Introduction, Methods, Results, Discussion. 
The concept has been applied to abstracts for a high-level overview of the four essential aspects of the work.
Structured abstracts~\cite{Nakayama2005Adoption} follows the IMRaD principles by including the same points in the abstract.
This motivated us to incorporate structure into automatic textual summaries, which can be easily parsed for the sake of information extraction (a.k.a. structured summary).

Figure~\ref{fig:approach} depicts a high-level view of the \textsc{MORTY} approach to information extraction on scholarly articles comprising several workflow phases.
It starts with pre-processing of the article text (i.e., the conversion from traditional PDF into text as well as cleaning and removing some needless segments of the text).
A summarization model is then capable of rendering a large text snippet into a much shorter structured summarization that contains pairs of properties and their corresponding values.
Later stages take care of parsing the produced summary via finding pre-defined syntactical patterns in the produced text.
Then interlinking extracted values to knowledge graph entities via exact lookup functionalities.
Lastly the newly extracted and aligned data gets added it to a destination knowledge graph.
The fundamental component of the approach is the summarization module due to the fact that all other components of \textsc{MORTY} are self-consistent.
This article tackles the following research question (\textbf{RQ}): \textit{How can we leverage structured summaries for the task of scholarly information extraction?}

\section{Evaluation}
\label{sec:eval}

Since the main component of \textsc{MORTY} is the structured summary generation, \textbf{we focus our evaluation on that component solely}.
Other components of the approach are deterministic in behavior and can be disregarded for the sake of this evaluation.
We created a dataset using the ORKG infrastructure, and empirically evaluated the feasibility of the summarization task with various models and approaches based on this dataset.

\subsection{Dataset Collection}
We require a source for human-curated annotations of scholarly articles.
ORKG is a knowledge graph that contains this sort of information.
Hence, we leverage the ORKG to create a dataset of scholarly articles' texts with a set of property-value pairs.
First, we took a snapshot of the ORKG data\footnote{Data snapshot was taken on 02.02.2022.} and we filtered on papers that are open access or have pre-prints on arXiv.
This ensures restriction-free access to the PDF files of articles.
Second, we parse the PDF files using GROBID~\cite{lopez2009grobid} into text.
Furthermore, we employ a heuristic to clean the text.
The heuristic involves the following steps:
i) Remove a set of pre-defined sections (such as abstract, related work, background, acknowledgments, and references);
ii) Remove all URLs from the text, as well as all Unicode characters;
iii) Remove tables, figures, footnotes, and citation texts.
Lastly, we collect all annotations from the ORKG excluding some properties that contain values of URIs and other structural properties\footnote{Properties that are used solely for information organization and have no semantic value.}.
Afterwards that data was collected in a format that the summarization model is trained on and can process. 
We split the data in 80-10-10 training-validation-testing split in favor of the testing set.

\begin{table}[]
\caption{Overview of models used in the evaluation, categorized per task. With the number of parameters, the max input size they can handle, and what dataset they are fine-tuned on beforehand to our training.}
\centering
\begin{tabular}{@{}l|ccc@{}}
\toprule
\multicolumn{1}{c|}{\textbf{Model}} & \textbf{\# of Params} & \textbf{Input Size} & \textbf{Finetuned on} \\ \midrule
\multicolumn{4}{c}{\textbf{Summarization}} \\ \midrule
\textbf{ProphetNet}-large~\cite{yan2020prophetnet}        & 391M & 2K & CNN      \\
\textbf{BART}-large~\cite{lewis2019bart}       & 460M & 4K  & CNN    \\
\textbf{GPT2}-large~\cite{radford2019language}       & 774M & 2K  & -      \\
\textbf{Pegasus}-large~\cite{zhang2020pegasus}   & 568M & 2K  & Pubmed \\
\textbf{BigBird}-large~\cite{zaheer2020big}    & 576M & 4K  & Pubmed \\
\textbf{T5}-large~\cite{raffel2019exploring}         & 770M & 4K  & -      \\
\textbf{Longformer}-large~\cite{beltagy2020longformer} & 459M & 8K  & Pubmed \\ \midrule
\multicolumn{4}{c}{\textbf{Question Answering}} \\ \midrule
\textbf{BERT}-large~\cite{devlin2018bert} & 335M & 1K  & SQuAD2 \\
\textbf{Longformer}-large~\cite{beltagy2020longformer} & 459M & 8K  & SQuAD2 \\ \midrule
\multicolumn{4}{c}{\textbf{Named Entity Recognition}} \\ \midrule
\textbf{BERT}-large~\cite{devlin2018bert} & 335M & 1K  & CoNLL \\ 
\textbf{RoBERTa}-large~\cite{liu2019roberta} & 355M & 2K  & CoNLL \\ \bottomrule
\end{tabular}
\label{tab:models}
\end{table}

\subsection{Baselines}
Throughout the evaluation, multiple baselines were investigated (see Table~\ref{tab:models}).
Various language models were used that are capable and pre-trained on summarization tasks. 
The maximum input size for each model varies depending on its architecture.
In our created dataset, the average entry contained around 5K tokens, with a maximum around 9K and a minimum around 1.5K.
Some of the models we used (e.g. Pegasus) are capable of abstractive summarization, i.e. can create summaries with words that don't exist in the original input text.
This is important when annotated properties and values are not present in the text, but are formulated differently.

Furthermore, the feasibility of the task is evaluated using two other categories of NLP tasks.
Extractive question answering (similarly to \cite{yao2014information}) language models are leveraged to try to extract values for certain questions.
The questions are formulated as follows: ``what is the \{property-label\}?''.
This type of baselines is inherently flawed because some of the properties and values from the datasets are not as is in the input text.
Another method for evaluation is to perform named entity recognition by recognizing the individual values as entities of interest and then classifying them into one of the classes (properties).

\subsection{Evaluation Results and Discussion}
The evaluation took place on a machine with 2 GPUs RTX A6000 each with a 48GB vRAM.
Training scripts where adapted from the fine-tuning scripts of each of model's code repositories with the help of the Transformers~\cite{wolf2020transformers} library.
The training used a batch size = 2 and epochs = 20 with early stopping enabled.

First, we evaluate the performance of various language models for structured summarization.
Table~\ref{tab:rouge} shows the results of the Rouge F1 metric (following~\cite{zhang2020pegasus}) for all considered language models.
Second, we evaluate the feasibility of the task using techniques other than summarization, namely extractive question answering and using named entity recognition.
Though the tasks of summarizations, named entitiy recognition, and question answering are not directly comparable;
we include this analysis to show the validity of summarization as a candidate for targeted IE tasks compared to other approaches.
Table~\ref{tab:accuracy} describes the performance of the two different approaches using two models for each case. 
Each model has different maximum input size and for the QA task the models were previously tuned on the SQuAD2~\cite{rajpurkar2018know} and the CoNLL~\cite{sang2003introduction} datasets for the NER task.
For the QA metrics, the reported number are computed @1, meaning only candidate results at the first place.

\begin{figure}[]
  \centering
  \includegraphics[width=\textwidth]{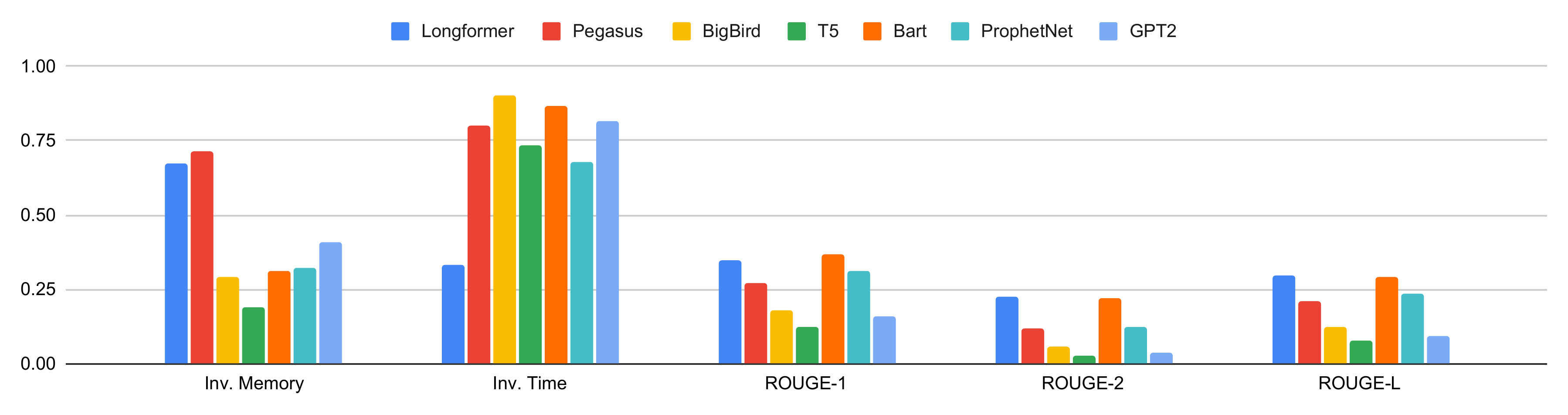}
  \caption{Summarization metrics overview of used models including the inverse time needed for training, and inverse memory consumption. Time and memory values are normalized and inversed. Higher values are better.}
  \label{fig:metrics}
\end{figure}

\begin{table}[]
\caption{Rouge F1 scores for 1-gram, 2-grams, and longest-gram variations of the summarization models. Top best results are indicated in bold, second best in italic.}
\centering
\resizebox{.7\textwidth}{!}{
\begin{tabular}{@{}l|ccc@{}}
\toprule
                    & \textbf{Rouge-1} & \textbf{Rouge-2} & \textbf{Rouge-L} \\ \midrule
\textbf{ProphetNet} & 31.1            & 12.5            & 23.7            \\
\textbf{BART}       & \textbf{36.7}   & \textit{22.0}   & \textit{29.4}   \\
\textbf{GPT2}       & 16.1            & 3.6            & 9.3            \\
\textbf{Pegasus}    & 27.1            & 11.7            & 21.2            \\
\textbf{BigBird}    & 17.9            & 5.9            & 12.5            \\
\textbf{T5}         & 12.2            & 2.8            & 7.9            \\
\textbf{Longformer} & \textit{34.7}   & \textbf{22.4}   & \textbf{29.6}   \\ \bottomrule
\end{tabular}
}
\label{tab:rouge}
\end{table}

\begin{table}[]
    \centering
    \caption{Precision, recall, f1-score results of other baseline models on the question answering (QA) and named entity recognition (NER) tasks.}
    \resizebox{.7\textwidth}{!}{
\begin{tabular}{@{}l|ccc@{}}
\toprule
                                         & \textbf{Precision} & \textbf{Recall} & \textbf{F1-score} \\ \midrule
\multicolumn{4}{c}{\textbf{Question Answering}} \\ \midrule
\multicolumn{1}{l|}{\textbf{BERT}}       &     20.8               &    18.1             &      19.1             \\
\multicolumn{1}{l|}{\textbf{Longformer}} &   23.7                  &   22.8              &  23.2                 \\ \midrule
\multicolumn{4}{c}{\textbf{Named Entity Recognition}} \\ \midrule
\multicolumn{1}{l|}{\textbf{BERT}}       &     17.2               &    17.0             &      17.0             \\
\multicolumn{1}{l|}{\textbf{RoBERTa}}    &     19.7               &    19.5             &      19.6             \\ \bottomrule
\end{tabular}
}
\label{tab:accuracy}
\end{table}

The results show that the task is viable using summarization and that structured summaries are able to extract the required information out of the scholarly articles.
When considering the normal summarization task, i.e. summarizing text into a coherent shorter text snippet, the top model~\cite{pang2022long} at the time of writing this article are performing with 51.05, 23.26, 46.47 for Rouge 1-2-L respectively\footnote{\url{https://paperswithcode.com/sota/text-summarization-on-pubmed-1}}.
This kind of summarization is far easier than structured summarization since the aim is merely coherent text creation, not structured summary of text fragments.
Examining Table~\ref{tab:rouge} and Table~\ref{tab:models}, we note that input size affects the performance of the model.
The summarization model requires the processing of the complete input article text to extract values from it, and if the model can not handle the full article then it will suffer in performance metrics. 
We note that, BART and Longformer summarization models performed best across all metrics.

ProphetNet and Pegaus performed well compared to other models, but they were not able to beat Longformer in part due to limited maximum input size.
GPT2 model suffered due to its nature as a generative model.
It was not able to generate the structured summary rather generating more coherent text.
Surprisingly, although big models with a 4K max input size BigBird and T5 did not perform comparatively to top models in the list.

Figure~\ref{fig:metrics} depicts an overview of the Rouge metrics of the summarization models as well as time and space requirement of each.
Though Longformer is the best performing model on average, it requires more time to train compared to BART. On the other hand, BART requires almost twice the memory compared to Longformer.

In order to empirically judge if the summarization method is suitable for the task of information extraction, we evaluate the approach against two categories of tasks: Question Answering (QA) and Named Entity Recognition (NER).
Table~\ref{tab:accuracy} shows the precision, recall, and f1-score metrics for two models in each category.
Due to the nature of the training data and the task itself, these two categories are inherently flawed because they are extractive and not abstractive, meaning that they aim at finding values from within the text, rather than compute with novel values.
Thus, these tasks are only able to retrieve parts of the values that are in the text and the rest are unattainable to them.
This explains why different models in both tasks preform poorly.

Table~\ref{tab:examples} shows some examples of five properties from three different articles with the expected values and the predicted values by the summarization model (here Longformer).
We observe that the model is able to extract partial values or similar values but with different wordings, as well as exact values, and completely different values.
For instance, ``Data size'' is an annotation property, were the expected value is not in the text, rather it is a summation of other values.
``Preprocessing steps'' property aggregate values from multiple places in the text.
The remarks made in this section answers our research question.

\begin{table}[t]
\caption{Examples: Expected vs. model predicted values.}
\centering
\resizebox{.85\textwidth}{!}{
\begin{tabular}{@{}ccc@{}}
\toprule
\textbf{Property}                        & \textbf{Expected} & \textbf{MORTY prediction} \\ \midrule
\multicolumn{1}{l|}{\textbf{Preprocessing steps}} &
  \begin{tabular}[c]{@{}c@{}}Topic segmentation\\ Anaphora resolution\\ Pronoun resolution\end{tabular} &
  Anaphora resolution \\ \midrule
\multicolumn{1}{l|}{\textbf{Data size}}  & 139 meetings      & 20 meetings               \\ \midrule
\multicolumn{1}{l|}{\textbf{Summarization type}}  & Abstractive      & Abstractive               \\ \midrule
\multicolumn{1}{l|}{\textbf{Evaluation metrics}}  & \begin{tabular}[c]{@{}c@{}}ROUGE-2\\ ROUGE-SU4 \end{tabular}     & F1               \\ \midrule
\multicolumn{1}{l|}{\textbf{Study location}} & Singapore         & The City of Singapore     \\ \bottomrule
\end{tabular}
}
\label{tab:examples}
\end{table}

\section{Conclusion and Future Directions}
\label{sec:conclusion}
The objective of this work was to leverage structured summarization for the task of IE from scholarly articles.
We evaluated various models on the summarization task, as well as compared against models performing question answering and named entity recognition.
The results show that summarization is a viable and feasible approach for the IE task on scholarly articles.
Based on our observations, we suggest the following open points in this domain:
i) Enable longer input sizes for large language models and evaluate them.
ii) Experiment with various structured summary formats and study their effect.
iii) Incorporate active learning with user feedback collected from a user interface within a scholarly infrastructure.
iv) Perform a user evaluation to study the efficacy of the IE task on scholarly data for users.

\bibliographystyle{splncs04}
\bibliography{references}
\end{document}